\documentclass[journal]{IEEEtran}

\usepackage[T1]{fontenc}
\usepackage[utf8]{inputenc}
\usepackage{color}
\usepackage[brazilian,english]{babel}

\usepackage{amsmath}
\usepackage{amssymb}
\usepackage{amsfonts}

\usepackage{graphicx}
\usepackage{psfrag}
\usepackage{subcaption}

\usepackage{amsthm}
\usepackage{thmtools}
\usepackage{thm-restate}

\usepackage{tikz}
\usepackage{pgfplots}
\usepgfplotslibrary{patchplots}
\usetikzlibrary{fit,positioning}

\usepackage{algpseudocode}

\usepackage{xspace}
\usepackage{multirow}
\usepackage{mathtools}
\usepackage{cancel}
\usepackage{footnote}

\newcommand{\R}{\mathbb{R}\xspace}
\newcommand{\X}{\mathcal{X}\xspace}

\theoremstyle{definition}
\newtheorem{definition}{Definition}

\begin{document}

\title{Hybrid Algorithm for Multi-Objective Optimization by Greedy Hypervolume
Maximization}
\author{Conrado S. Miranda,
Fernando J. Von Zuben,~\IEEEmembership{Senior Member,~IEEE}
\thanks{C. S. Miranda and F. J. Von Zuben are with the School of Electrical and
  Computer Engineering, University of Campinas (Unicamp), Brazil. E-mail:
\{conrado,vonzuben\}@dca.fee.unicamp.br}}

\maketitle

\begin{abstract}
  This paper introduces a high-performance hybrid algorithm, called Hybrid
  Hypervolume Maximization Algorithm (H2MA), for multi-objective optimization
  that alternates between exploring the decision space and exploiting the
  already obtained non-dominated solutions. The proposal is centered on
  maximizing the hypervolume indicator, thus converting the multi-objective
  problem into a single-objective one. The exploitation employs gradient-based
  methods, but considering a single candidate efficient solution at a time, to
  overcome limitations associated with population-based approaches and also to
  allow an easy control of the number of solutions provided. There is an
  interchange between two steps. The first step is a deterministic local
  exploration, endowed with an automatic procedure to detect stagnation. When
  stagnation is detected, the search is switched to a second step characterized
  by a stochastic global exploration using an evolutionary algorithm. Using five
  ZDT benchmarks with 30 variables, the performance of the new algorithm is
  compared to state-of-the-art algorithms for multi-objective optimization, more
  specifically NSGA-II, SPEA2, and SMS-EMOA. The solutions found by the H2MA
  guide to higher hypervolume and smaller distance to the true Pareto frontier
  with significantly less function evaluations, even when the gradient is
  estimated numerically. Furthermore, although only continuous decision spaces
  have been considered here, discrete decision spaces could also have been
  treated, replacing gradient-based search by hill-climbing. Finally, a thorough
  explanation is provided to support the expressive gain in performance that was
  achieved.
\end{abstract}

\begin{IEEEkeywords}
  Exploration-exploitation algorithm;
  Gradient-based optimization;
  Hypervolume maximization;
  Multi-objective optimization.
\end{IEEEkeywords}

\IEEEpeerreviewmaketitle

\section{Introduction}
\IEEEPARstart{M}{ulti-objective} optimization (MOO) is a generalization of the
standard single-objective optimization to problems where multiple criteria are
defined and they conflict with each other~\cite{miettinen1999nonlinear}. In this
case, there can be multiple optimal solutions with different trade-offs between
the objectives. Since the optimal set can be continuous, an MOO problem is given
by finding samples from the optimal set, called Pareto set. However, we also
wish that the projection of the obtained samples of the Pareto set into the
objective space be well-distributed along the Pareto frontier, which is the
counterpart for the Pareto set, so that the solutions present more diverse
trade-offs.

The current state-of-the-art for MOO relies on the use of evolutionary
algorithms for finding the desired samples~\cite{deb2001multi}. One of these
algorithms is the NSGA-II~\cite{deb2002fast}, which performs non-dominance
sorting, thus ordering the proposed solutions according to their relative
dominance degree, and dividing the solution set in subsequent frontiers of
non-dominated solutions. NSGA-II also uses crowding distance, which measures how
close the nearby solutions are, to maintain diversity in the objective space.
Another well-known algorithm is the SPEA2~\cite{zitzler2001spea2}, where the
solutions have a selective pressure to move towards the Pareto frontier and also
to stay away from each other.

These algorithms are based on heuristics to define what characterizes a good set
of solutions. However, the hypervolume indicator~\cite{zitzler2007hypervolume}
defines a metric of performance for a set of solutions, thus allowing a direct
comparison of multiple distinct sets of solutions~\cite{zitzler2003performance},
with higher values indicating possible better quality. The hypervolume is
maximal at the Pareto frontier and increases if the samples are better
distributed along the frontier~\cite{auger2009theory}. Due to these properties,
it represents a good candidate to be maximized in MOO, being explicitly explored
in the SMS-EMOA~\cite{beume2007sms}, where solutions that contribute the least
to the hypervolume are discarded.

On the other hand, local search methods have been successful in single-objective
optimization (SOO) due to their efficiency in finding a local optimum for some
problems~\cite{bertsekas1999nonlinear,luenberger2008linear}, so that research
has been performed to try to adapt these methods for MOO problems. For instance,
\cite{bosman2012gradients} defined a method for finding all minimizing
directions in a MOO problem, but the proposed algorithm achieved low performance
on usual benchmark functions.

Alternatively, instead of adapting the single-objective methods to work on MOO
problems, we can create a SOO problem associated with the MOO one, such that a
good solution for the single-objective case is a good solution for the
multi-objective case. Since the hypervolume is able to describe how good a
population is, based on a single indicator, the MOO problem can be converted
into the maximization of the population's hypervolume.

Based on this idea, \cite{hypervolume_gradient1} proposed a method to compute
the hypervolume's gradient for a given population, so that the optimal search
direction for each individual could be established. However,
\cite{hypervolume_gradient2} showed that adjusting the population through
integration of the hypervolume's gradient not always work, with some initially
non-dominated points becoming dominated and others changing very little over the
integration.

In this paper, we introduce an algorithm for maximizing the hypervolume by
optimizing one point at a time, instead of adjusting a whole population at once.
The algorithm alternates between exploring the space for non-dominated solutions
and, when they are found, exploiting them using local search methods to maximize
the populations' hypervolume when only this active point can be moved.
Therefore, once the hypervolume has converged, which is guaranteed to happen
because the problem is bounded, the point is fixed in all further iterations. We
found that this restriction is enough to overcome the issues presented
in~\cite{hypervolume_gradient2} when using the hypervolume's gradient. The
proposed algorithm, called Hybrid Hypervolume Maximization Algorithm (H2MA), is
a hybrid one, since it is composed of global exploration and local exploitation
procedures, properly managed to be executed alternately.

Results over the ZDT benchmark~\cite{zdt2000a} show that the new algorithm
performs better than the state-of-the-art evolutionary algorithms, both in terms
of total hypervolume and distance to the Pareto frontier. Moreover, the
algorithm was able to work deterministically in most of the benchmark problems,
which makes it less susceptible to variations due to random number generation.
Due to the high quality of the solutions found in less function evaluations than
what is achieved by the current state-of-the-art, we consider that the new
algorithm is a viable choice for solving MOO problems. Moreover, since a single
solution is introduced at a time, the user is able to stop the algorithm when
the desired number of solutions is found, while evolutionary algorithms must
evolve the whole population at the same time.

This paper is organized as follows. Section~\ref{sec:moo} introduces the
concepts of multi-objective optimization required, including the hypervolume
indicator, and discusses the problems with the gradient-based approach for
hypervolume maximization introduced in~\cite{hypervolume_gradient2}.
Section~\ref{sec:hybrid} provides the details of the new H2MA algorithm, and
Section~\ref{sec:results} shows the comparison with the state-of-the-art
algorithms. Finally, Section~\ref{sec:conclusion} summarizes the results and
discusses future research direction.
\section{Multi-Objective Optimization and the Hypervolume Indicator}
\label{sec:moo}
A multi-objective optimization problem is described by its decision space $\X$
and a set of objective functions $f_i(x) \colon \X \to \mathcal Y_i, i \in
\{1,\ldots,M\}$, where $\mathcal Y_i \subseteq \R$ is the associated objective
space for each objective function~\cite{deb2014multi}. Due to the symmetry
between maximization and minimization, only the minimization problem is
considered here. Each point $x$ in the decision space has a counterpart in the
objective space $\mathcal Y = \mathcal Y_1 \times \cdots \times \mathcal Y_M$
given by $y = f(x) = (f_1(x),\ldots,f_M(x))$.

Since there are multiple objectives, a new operator for comparing solutions must
be used, since the conventional ``less than'' operator $<$ can only compare two
numbers. This operator is denoted the dominance operator and is defined as
follows.
\begin{definition}[Dominance]
  \label{def:dominance}
  Let $y$ and $y'$ be points in $\mathcal Y$, the objective space. Then $y$
  dominates $y'$, denoted $y \prec y'$, if $y_i < y'_i$ for all $i$.
\end{definition}

From this definition, a point $y$ that dominates another point $y'$ is better
than $y'$ in all objectives. Thus, there is no reason someone would choose $y'$
over $y$, and it can be discarded, as occurs in many multi-objective
optimization algorithms~\cite{deb2014multi}. Note that there are other
definitions of the dominance operator~\cite{zitzler2003performance}, where one
considers the inequality $\le$ instead of the strict inequality $<$ used here.
However, equality in some of the coordinates may be an issue when using the
hypervolume indicator, such as when taking its
derivative~\cite{hypervolume_gradient1}. This is why the strict version is used
in this work.

Using the dominance, we can define the set of points characterized by the fact
that no other point can have better performance in all objectives.

\begin{definition}[Pareto Set and Frontier]
  The Pareto set is defined by the set of all points in the decision space that
  are not dominated by any other point in the decision space, when using the
  objectives. That is, the Pareto set is given by $\mathcal P = \{x \in \X \mid
  \nexists x' \in \X \colon f(x') \prec f(x)\}$. The Pareto frontier is the
  associated set in the objective space, given by $\mathcal F = \{f(x) \mid x
  \in \mathcal P\}$.
\end{definition}

\subsection{The Hypervolume Indicator}
In order to define the hypervolume indicator~\cite{zitzler2007hypervolume}, we
must first define the Nadir point, which is a point in the objective space that
is dominated by every point in a set.
\begin{definition}[Nadir Point]
  \label{def:nadir}
  Let $X = \{x_1,\ldots,x_N\} \in \X^N$ be a set of points in the decision
  space. Let $z \in \R^M$ be a point in the objective space. Then $z$ is a valid
  Nadir point if, for all $x \in X$ and $i \in \{1,\ldots,M\}$, we have that
  $f_i(x) < z_i$. Using Definition~\ref{def:dominance}, this can be written as
  $f(x) \prec z$.
\end{definition}

Again, it is possible to allow equality in the definition of the Nadir point,
just like in the definition of dominance. However, when equality is allowed, it
is possible for some point to have a null hypervolume, which can guide to
undesired decisions when using the hypervolume as a performance metric, since
such points would not contribute to the hypervolume and would be replaced by
other points. Using the definition of a Nadir point, we can define the
hypervolume for a set of points.

\begin{definition}[Hypervolume]
  \label{def:hypervolume}
  Let $X = \{x_1,\ldots,x_N\} \in \X^N$ be a set of points in the decision
  space. Let $z \in \R^M$ be a valid Nadir point in the objective space. Then
  the hypervolume can be defined as:
  \begin{equation}
    \label{eq:hypervolume}
    H(X;z) = \int_{\R^M} 1[\exists x \in X \colon f(x) \prec y \prec
      z] \text{d}y,
  \end{equation}
  where $1[\cdot]$ is the indicator function.
\end{definition}

\begin{figure}[t]
  \centering
  \psfrag{y1}[c][b]{$y_1$}
  \psfrag{y2}[c][t]{$y_2$}
  \includegraphics[width=0.6\linewidth]{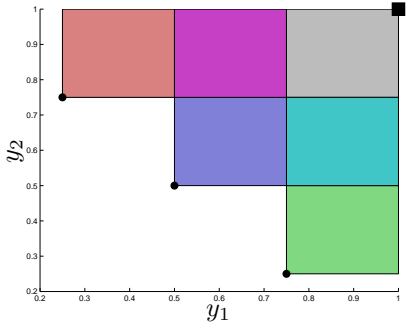}
  \caption{Example of hypervolume. The non-dominated solutions in the objective
    space are shown in black circles, and the Nadir point is shown in the black
    square. For each non-dominated solution, the region between it and the Nadir
    point is filled, with colors combining when there is overlap, and the total
  hypervolume is given by the area of the shaded regions. Best viewed in color.}
  \label{fig:hypervolume_example}
\end{figure}

The hypervolume measures how much of the objective space is dominated by a
current set $X$ and dominates the Nadir point $z$.
Fig.~\ref{fig:hypervolume_example} shows an example of the hypervolume for a set
of three non-dominated points. For each point, the shaded region represents the
area dominated by the given point, with colors combining when there is overlap.

\subsection{Gradient of the Hypervolume}
\label{sec:moo:gradient}
As stated earlier, since the hypervolume provides such a good indicator of
performance in multi-objective problems, it can be used to transform the
multi-objective problem into a single-objective one, characterized by the
maximization of the hypervolume.

Although such approach proved to be successful when using evolutionary
algorithms as the optimization method~\cite{beume2007sms}, the same did not
happen when using the hypervolume's gradient to perform the
optimization~\cite{hypervolume_gradient2}. However, it is well-known that
gradient methods have been successful in single-objective
optimization~\cite{bertsekas1999nonlinear,luenberger2008linear}, thus suggesting
that they should be a reasonable choice for multi-objective optimization devoted
to maximizing the hypervolume, since the hypervolume operator is well-defined
almost everywhere in the objective space.

The hypervolume's gradient for a set of points was introduced
in~\cite{hypervolume_gradient1}, and it can be used to compute the optimal
direction in which a given point should move to increase the hypervolume
associated with the current set of non-dominated solutions. Although the
hypervolume is not a continuously differentiable function of its arguments,
since dominated points do not contribute to the hypervolume and thus have null
gradient, the gradient can be computed whenever any two points have different
values for all objectives.

Based on this motivation, \cite{hypervolume_gradient2} used the hypervolume's
gradient as a guide for adjusting a set of points by numerical integration, that
is, performing a small step in the direction pointed by the gradient. Even
though the algorithm was able to achieve the Pareto set in some cases, it failed
to converge to efficient points when some points got stuck along the iterative
process, either because their gradients became very small or because they became
dominated by other points. Once dominated, these points do not contribute to the
hypervolume and remain fixed. This causes a major issue to using the hypervolume
gradient in practice, since dominated points can be discarded, because there is
no possibility to revert them to non-dominated points anymore, and the points
with small gradients remain almost stagnant.

If we analyze Eq.~\eqref{eq:hypervolume}, we can see that points at the border
in the objective space are the only ones that can fill some portions of the
objective space. On the other hand, points that are not at the border have less
influence in the hypervolume, since part of the area dominated by them is also
dominated by some other points. In the analysis presented
in~\cite{hypervolume_gradient2}, it is clear that the cases where some points
got stuck had higher gradients for the border points in the objective space,
which led to the dominance or decrease of contribution of some or all central
points.

To make this idea clearer, consider the example in
Fig.~\ref{fig:hypervolume_example}. If the point located at $(0.75,0.25)$
decreases its value on the second objective, it can increase the population's
hypervolume. Moreover, it is the only point that can do so without competition
for that portion of the space, since it is the point with the largest value for
the first objective. The same holds for the point at $(0.25,0.75)$ and the first
objective.

However the point located at $(0.5,0.5)$ has to compete with the other two
points to be the sole contributor for some regions. Therefore, its effect on the
hypervolume is smaller, which leads to a smaller gradient. Furthermore, if less
area is dominated by the middle point alone, which can occur during the points
adjustment as the middle one moves less, then its influence becomes even smaller
and it can become dominated.

It is important to highlight that this behavior does not happen always, but can
occur along the iterative process, as shown in~\cite{hypervolume_gradient2}.
This leads to the base hypothesis for the algorithm developed in this paper:
when using the hypervolume's gradient for optimization, the competition for
increasing the hypervolume among points should be avoided.
\section{Hybrid Hypervolume Maximization Algorithm}
\label{sec:hybrid}
From the discussion in Section~\ref{sec:moo:gradient}, one can see that the
major problem when optimizing the hypervolume directly using its gradient may be
the competition among points. Therefore, our proposed algorithm optimizes a
single solution at a time, avoiding this competition.

Theoretically, the algorithm can be described by choosing a new point that
maximizes the hypervolume when taking into account the previous points, such
that its recurring equation can be written as:
\begin{equation}
  \label{eq:greedy}
  x_t = \arg \max_{x \in \X} H(X_{t-1} \cup \{x\}),
  X_t = X_{t-1} \cup \{x_t\}, t \in \mathbb N,
\end{equation}
where the initial set is given by $X_{-1} = \{\}$.

Since a single point is being optimized at a time, the optimization becomes
simpler and, as we will show in Section~\ref{sec:results}, requires less
function evaluations. Moreover, one could argue that maintaining the previous
set fixed reduces the flexibility allowed in comparison with a set where all the
points are being concurrently adjusted. Although this may be true, we will also
show in Section~\ref{sec:results} that the proposed algorithm performs well
despite this loss of flexibility.

\begin{figure}[t]
  \small
  \begin{algorithmic}
    \Require{Objectives $f$}
    \Require{Design space $\X$}
    \Require{Nadir point $z$}
    \Ensure{Set of candidate solutions $X$}
    \Function{HybridGreedyOptimizer}{$f, \X, z$}
      \State{$Regions,X$ $\gets$ \Call{CreateInitialRegion}{$f, \X$}}
      \While{\textbf{not} \emph{stop condition} \textbf{and} $|Regions| > 0$}
        \State{$R \gets Regions$.pop()}
        \Comment{Removes the region with the largest volume}
        \State{$x_0 \gets$ \Call{ExploreDeterministic}{$f,\X,R,X$}}
        \If{$x_0$ is valid}
          \State{$x \gets$ \Call{Exploit}{$f,\X,x_0,X,z$}}
          \State{$NewRegions \gets$ \Call{CreateRegions}{$R,x,f$}}
          \State{$Regions \gets Regions \cup NewRegions$}
          \State{$X \gets X \cup \{x\}$}
        \EndIf
      \EndWhile
      \While{\textbf{not} \emph{stop condition}}
        \State{$x_0 \gets$ \Call{ExploreStochastic}{$f,\X,X$}}
        \State{$x \gets$ \Call{Exploit}{$f,\X,x_0,X,z$}}
        \State{$X \gets X \cup \{x\}$}
      \EndWhile
      \State{\Return{$X$}}
    \EndFunction
  \end{algorithmic}
  \caption{Hybrid algorithm that performs deterministic and stochastic
  exploration until a suitable solution is found, and then exploits it.}
  \label{alg:main}
\end{figure}
\begin{figure}[t]
  \small
  \begin{algorithmic}
    \Require{Objectives $f$}
    \Require{Design space $\X$}
    \Require{Current exploration region $R$}
    \Require{Set of candidate solutions $X$}
    \Ensure{New initial condition $x_0$}
    \Function{ExploreDeterministic}{$f,\X,R,X$}
      \State{$x_0 \gets$ \Call{Mean}{$R.X$}}
      \State{Minimize $\|R.mid - f(x)\|$ from $x_0$ until some candidate $x$ is
      not dominated by $X$}
      \If{found non-dominated $x$}
        \State{$x_0 \gets x$}
      \Else
        \State{$x_0 \gets$ some invalid state}
      \EndIf
      \State{\Return{$x_0$}}
    \EndFunction
  \end{algorithmic}
  \caption{A deterministic exploration is performed based on some region.}
  \label{alg:exploration}
\end{figure}
\begin{figure}[t]
  \small
  \begin{algorithmic}
    \Require{Objectives $f$}
    \Require{Design space $\X$}
    \Ensure{Set of candidate solutions $X$}
    \Ensure{Initial exploration region $R$}
    \Function{CreateInitialRegion}{$f, \X$}
      \State{$X \gets \{\}$}
      \State{$x_0 \gets \X.mean$}
      \Comment{Gets the average candidate}
      \For{$i=1,\ldots,|f|$}
        \State{$x \gets$ \Call{Minimize}{$f_i,x_0,\X$}}
        \State{$X \gets X \cup \{x\}$}
      \EndFor
      \State{$R \gets$ \Call{CreateRegion}{$X,f$}}
      \State{\Return{$\{R\},X$}}
    \EndFunction
  \end{algorithmic}
  \caption{The initial region is created from the points that minimize each
  objective individually.}
  \label{alg:initial}
\end{figure}

The algorithm described in Eq.~\eqref{eq:greedy} is theoretically ideal, since
finding the maximum is hard in practice. Therefore, the actual algorithm
proposed is shown in Fig.~\ref{alg:main}. This algorithm performs exploration of
the objective space until a new solution that is not dominated by the previous
candidate solutions is found. When it happens, the hypervolume of the whole
set is larger than the hypervolume when considering only previous candidate
solutions.

The new candidate solution is then exploited to maximize the total hypervolume
and, after convergence, is then added to the existing set. It is important to
highlight that the exploitation phase cannot make the solution become dominated,
since that would reduce the hypervolume in comparison with the initial
condition. Therefore, the problem of points becoming dominated during the
exploitation is avoided. Furthermore, the exploitation is a traditional
single-objective optimization, so that gradient methods can be used if the
decision set $\X$ is continuous or hill-climbing methods can be used for
discrete $\X$.

Once finished the exploitation, the algorithm begins the exploration phase
again. The exploration can be deterministic, based on regions of the objective
space defined by previous solutions, or stochastic, where a stochastic
algorithm, such as an evolutionary algorithm, is used to find the new candidate.
When a non-dominated candidate is found, the algorithm turns to exploitation
again.

We highlight that the deterministic exploitation algorithm proposed is based on
the definition of these regions, but other deterministic methods can be used.
However, the algorithm must be able to establish when it is not able to provide
further improvements, so that the change to the stochastic global exploration
can be made. In the algorithm shown in Fig.~\ref{alg:main}, regions that do not
provide a valid initial condition are discarded without creating new regions, so
that eventually the algorithm can switch to the stochastic global exploration.

The algorithm for deterministic exploration is shown in
Fig.~\ref{alg:exploration}. It combines the points used to create a given region
in order to produce an initial condition and tries to minimize the distance
between its objective value and a reference point. Once a non-dominated point is
found, it is returned for exploitation. Although this simple optimization
provided good results without requiring many function evaluations, other methods
can be used to perform this exploration. Alternatively, one can also perform a
stochastic exploration instead of a deterministic one, but this may have
negative effects on the performance if the information provided by the output
(region R) is not used, since a global search would be required.

\begin{figure}[t]
  \small
  \begin{algorithmic}
    \Require{Current explored region $R$}
    \Require{Current solution $x$}
    \Require{Objectives $f$}
    \Ensure{New exploration regions $NewRegions$}
    \Function{CreateRegions}{$R,x,f$}
      \State{$NewRegions \gets \{\}$}
      \For{$X'$ \textbf{in} \Call{Combinations}{$R.X,|R.X|-1$}}
        \State{$R' \gets$ \Call{CreateRegion}{$X' \cup \{x\},f$}}
        \State{$NewRegions \gets NewRegions \cup \{R'\}$}
      \EndFor
      \State{\Return{$NewRegions$}}
    \EndFunction
  \end{algorithmic}
  \caption{New exploration regions are created by combining the current solution
  with the previous region.}
  \label{alg:regions}
\end{figure}
\begin{figure}[t]
  \small
  \begin{algorithmic}
    \Require{Objectives $f$}
    \Require{Set of candidate solutions $X$}
    \Ensure{Exploration region $R$}
    \Function{CreateRegion}{$X,f$}
      \State{$V = \prod_{i=1}^{|f|} (\max_{x \in X} f_i(x) - \min_{x \in X}
      f_i(x))$}
      \If{$V > 0$}
        \State{$R.X \gets X$}
        \State{$R.mid \gets$ \Call{Mean}{$\{f(x) \mid x \in X\}$}}
        \State{$R.V \gets V$}
      \Else
        \State{$R \gets$ null element such that $\{R\} \equiv \{\}$}
      \EndIf
      \State{\Return{$R$}}
    \EndFunction
  \end{algorithmic}
  \caption{An exploration region is created from a set of candidates if the
  region have some volume.}
  \label{alg:region}
\end{figure}

The first region is created by finding points that minimize each objective
separately, as shown in Fig.~\ref{alg:initial}. This establishes that the
initial region will have a number of candidate solutions associated with it
equal to the number of objectives, so that the solutions are at the border of
the region.

\begin{figure*}[t]
  \centering
  \begin{subfigure}[b]{0.4\linewidth}
    \psfrag{p1}[c][c]{$f(x_1)$}
    \psfrag{p2}[c][c]{$f(x_2)$}
    \psfrag{p3}[c][c]{$f(\overline x_{12})$}
    \psfrag{p4}[c][c]{$f(x')$}
    \psfrag{p5}[c][c]{$\overline y_{12}$}
    \psfrag{y1}[c][b]{$f_1(\cdot)$}
    \psfrag{y2}[c][t]{$f_2(\cdot)$}
    \psfrag{R}[c][c]{$R$}
    \includegraphics[width=\linewidth]{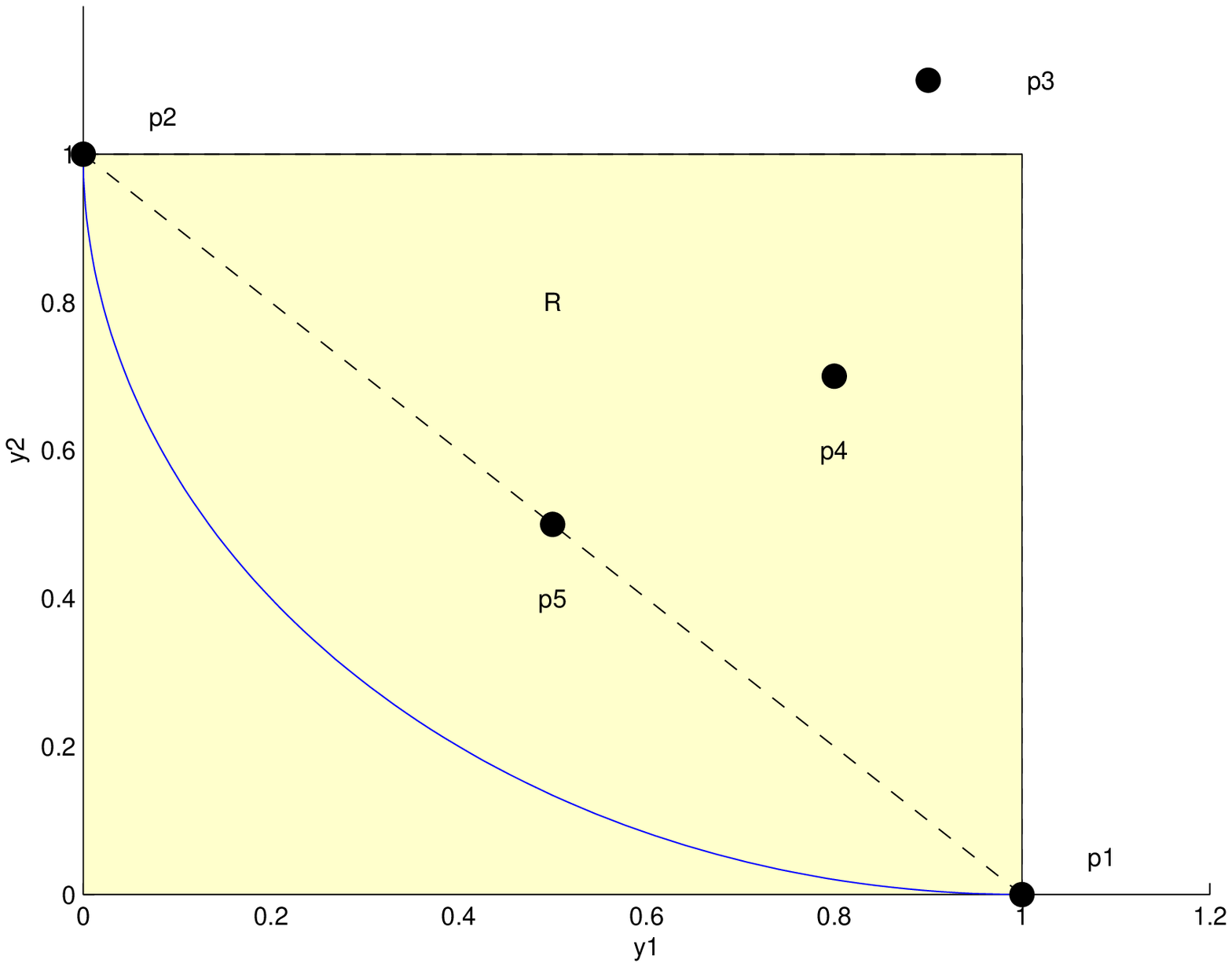}
    \caption{Deterministic exploration}
    \label{fig:algorithm_example:exploration}
  \end{subfigure}
  \quad
  \begin{subfigure}[b]{0.4\linewidth}
    \psfrag{p1}[c][c]{$f(x_1)$}
    \psfrag{p2}[c][c]{$f(x_2)$}
    \psfrag{p4}[c][c]{$f(x')$}
    \psfrag{p6}[c][c]{$f(x^*)$}
    \psfrag{h1}[c][c]{$H^*$}
    \psfrag{h2}[c][c]{$H'$}
    \psfrag{y1}[c][b]{$f_1(\cdot)$}
    \psfrag{y2}[c][t]{$f_2(\cdot)$}
    \psfrag{R2}[c][c]{$R_1$}
    \psfrag{R1}[c][c]{$R_2$}
    \includegraphics[width=\linewidth]{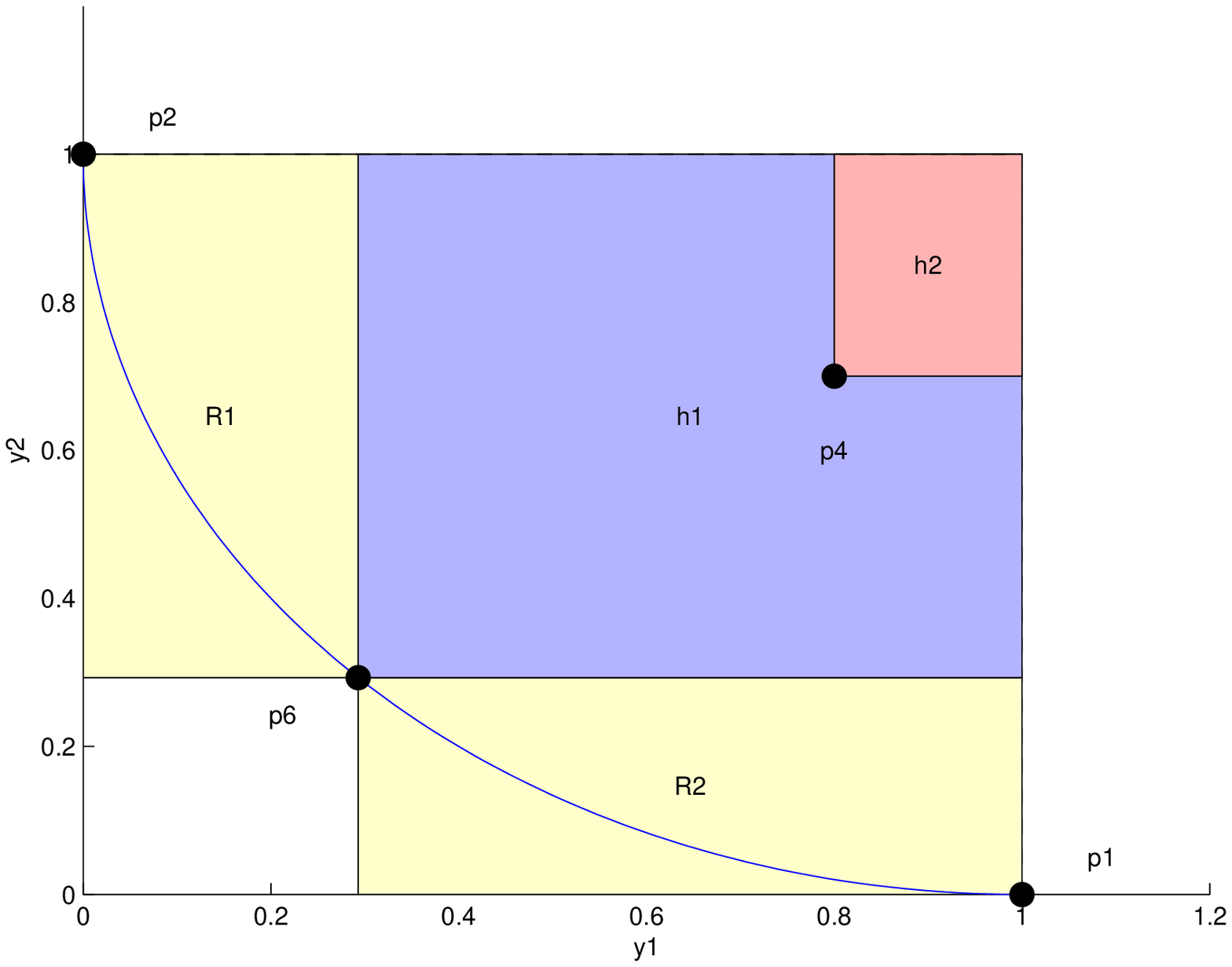}
    \caption{Exploitation}
    \label{fig:algorithm_example:exploitation}
  \end{subfigure}
  \caption{Deterministic exploration and exploitation steps of the new algorithm
    in an example problem. The Pareto frontier is shown in the blue line, and
  the regions used by the deterministic exploration are shown in yellow.}
  \label{fig:algorithm_example}
\end{figure*}

When new regions are created after exploitation, we ignore the solutions that
created the region, one at a time, and replace it with the proposed new
solution, as shown in Fig.~\ref{alg:regions}, to create a new region. This
guarantees that the number of solutions for each region is kept equal to the
number of objectives.

Finally, Fig.~\ref{alg:region} shows how a region is created. If a region does
not have a volume, then at least one objective for two solutions is the same.
Although we could allow such region to exist without modifying the rest of the
algorithm, these regions tend to not provide good candidates for exploitation
and delay the change to stochastic global exploration. Furthermore, one can even
prohibit regions with volume smaller than some known constant, as they probably
will not provide good exploitation points, and the change to stochastic global
exploration happens earlier.

Fig.~\ref{fig:algorithm_example} shows a step of the algorithm in an example
problem with two objectives.
The deterministic exploration receives a
region $R$, composed of the points $x_1$ and $x_2$. The mean of the points that
compose the region is given by $\overline x_{12} = (x_1+x_2)/2$ and its
evaluation in the objective space is shown in
Fig.~\ref{fig:algorithm_example:exploration}. The mean objective of the points
that compose the region is also computed and is shown as $\overline y_{12} =
(f(x_1)+f(x_2))/2$. The deterministic exploration is then defined by the
problem
\begin{equation}
  \label{eq:deterministc_exploration}
  \min_{x \in \X} \|f(x) - \overline y_{12}\|,
\end{equation}
which uses $\overline x_{12}$ as the initial condition for the optimization.
Since $\overline y_{12}$ is guaranteed to be non-dominated by $f(x_1)$ and
$f(x_2)$, this should guide the search to the non-dominated region of the space.

While performing this optimization, some intermediary points are evaluated,
either while computing the numeric gradient or after performing a gradient
descent step. The deterministic exploration stops as soon as a non-dominated
point is found, which is given by $f(x')$ in the example in
Fig.~\ref{fig:algorithm_example:exploration}. Note that this example shows
$f(\overline x_{12})$ as being dominated by $f(x_1)$ and $f(x_2)$, but it can
also be non-dominated. In this case, $x' = \overline x_{12}$ and no optimization
step for the problem in Eq.~\eqref{eq:deterministc_exploration} is performed.
Supposing no non-dominated point $f(x')$ is found during the deterministic
exploration, the region is simply discarded, without performing an exploitation
step.

Using the point $x'$, whose $f(x')$ is non-dominated, provided by the
deterministic or stochastic exploration, the exploitation is performed.
Fig.~\ref{fig:algorithm_example:exploitation} shows the hypervolume
contributions for the initial point $x'$ and the optimal point $x^*$, which
maximizes the total hypervolume as in Eq.~\eqref{eq:greedy}. Since $x'$ is
non-dominated, its hypervolume contribution $H'$ is positive and the hypervolume
gradient relative to the objectives is non-zero. After finding $x^*$ and if $x'$
was provided by the deterministic exploration, new regions must be created
to allow further exploration. Therefore, according to Fig.~\ref{alg:regions},
the regions $R_1 = (x_1, x^*)$ and $R_2 = (x_2, x^*)$ are created for further
exploration.

This finalizes a step of the algorithm, which is repeated until the
given stop condition is not met. As at most one point is found by each step, the
stop condition can be defined based on the number of desired points.

Note that all the methods used in this algorithm assume that the optimization,
either for exploitation or for minimizing one objective alone, requires an
initial condition. This is true for hill climbing or gradient methods, but the
algorithm can easily be modified if the optimization does not require it.
\section{Experimental Results}
\label{sec:results}

\begin{figure*}[t]
  \centering
  \begin{subfigure}[b]{0.48\linewidth}
    \psfrag{evals}[c][b]{\# evaluations}
    \psfrag{H}[c][c]{H}
    \includegraphics[width=\linewidth]{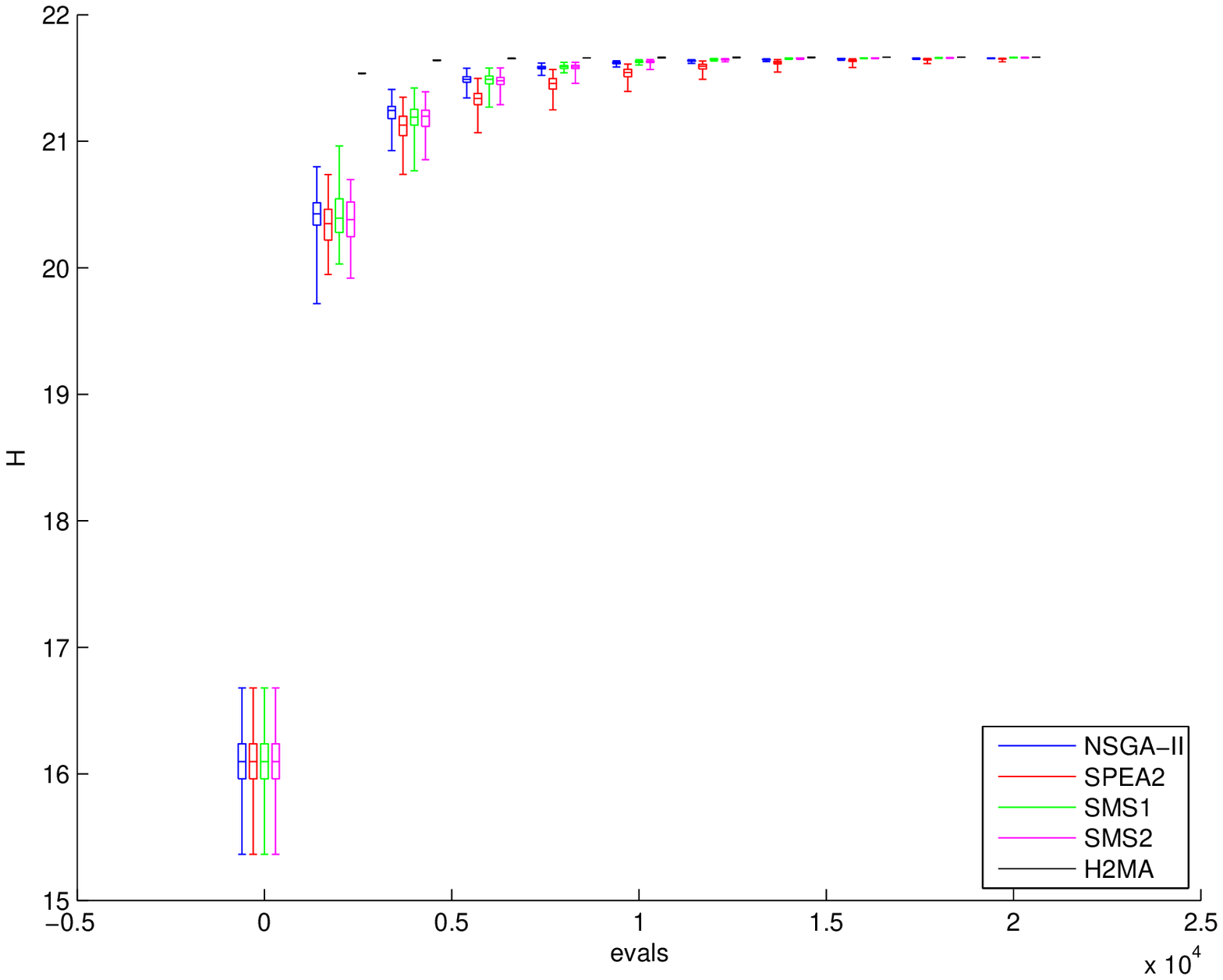}
    \caption{Hypervolume}
    \label{fig:zdt1:H}
  \end{subfigure}
  \begin{subfigure}[b]{0.48\linewidth}
    \psfrag{evals}[c][b]{\# evaluations}
    \psfrag{P}[c][c]{$\log_{10}$ P}
    \includegraphics[width=\linewidth]{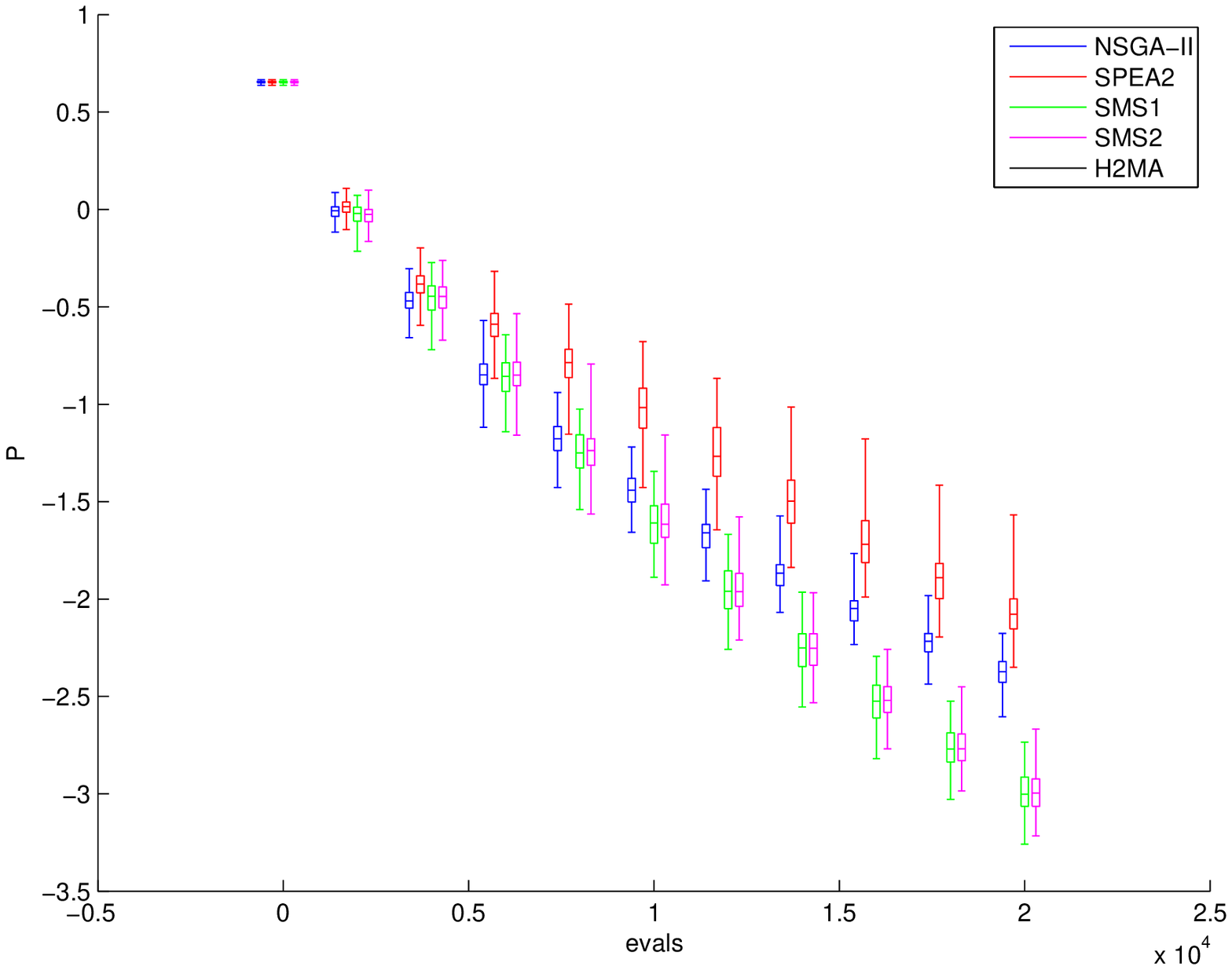}
    \caption{P-distance. Zero values not shown.}
    \label{fig:zdt1:P}
  \end{subfigure}
  \caption{0th, 25th, 50th, 75th, and 100th percentiles every 2000 evaluations
  for the all algorithms on ZDT1.}
  \label{fig:zdt1}
\end{figure*}

To compare the algorithm proposed in Section~\ref{sec:hybrid}, called Hybrid
Hypervolume Maximization Algorithm (H2MA), with the existing algorithms, the ZDT
family of functions~\cite{zdt2000a} was chosen. These functions define a common
benchmark set in the multi-objective optimization literature, since they define
a wide range of problems to test different characteristics of the optimization
algorithm. All functions defined in~\cite{zdt2000a} have a continuous decision
space $\X$, except for the ZDT5 which has a binary space. In this paper, only
the continuous test functions were used to evaluate the performance of the new
algorithm, and their equations are shown in the appendix.

Table~\ref{tab:experiment} provides a summary of the evaluation functions, their
decision spaces, and the Nadir points used to compute the hypervolume. The Nadir
points are defined by upper bounds of the objectives, which guarantees that the
hypervolume computation is always valid, plus one, since not adding an extra
value would mean that points at the border of the frontier would have no
contribution to the hypervolume and would be avoided. In all instances, a total
of $n=30$ variables were considered, as common in the literature. The
evolutionary algorithms' and evaluation functions' implementations were given by
the PaGMO library~\cite{biscani2010global}.

\begin{figure*}[t]
  \centering
  \begin{subfigure}[b]{0.48\linewidth}
    \psfrag{evals}[c][b]{\# evaluations}
    \psfrag{H}[c][c]{H}
    \includegraphics[width=\linewidth]{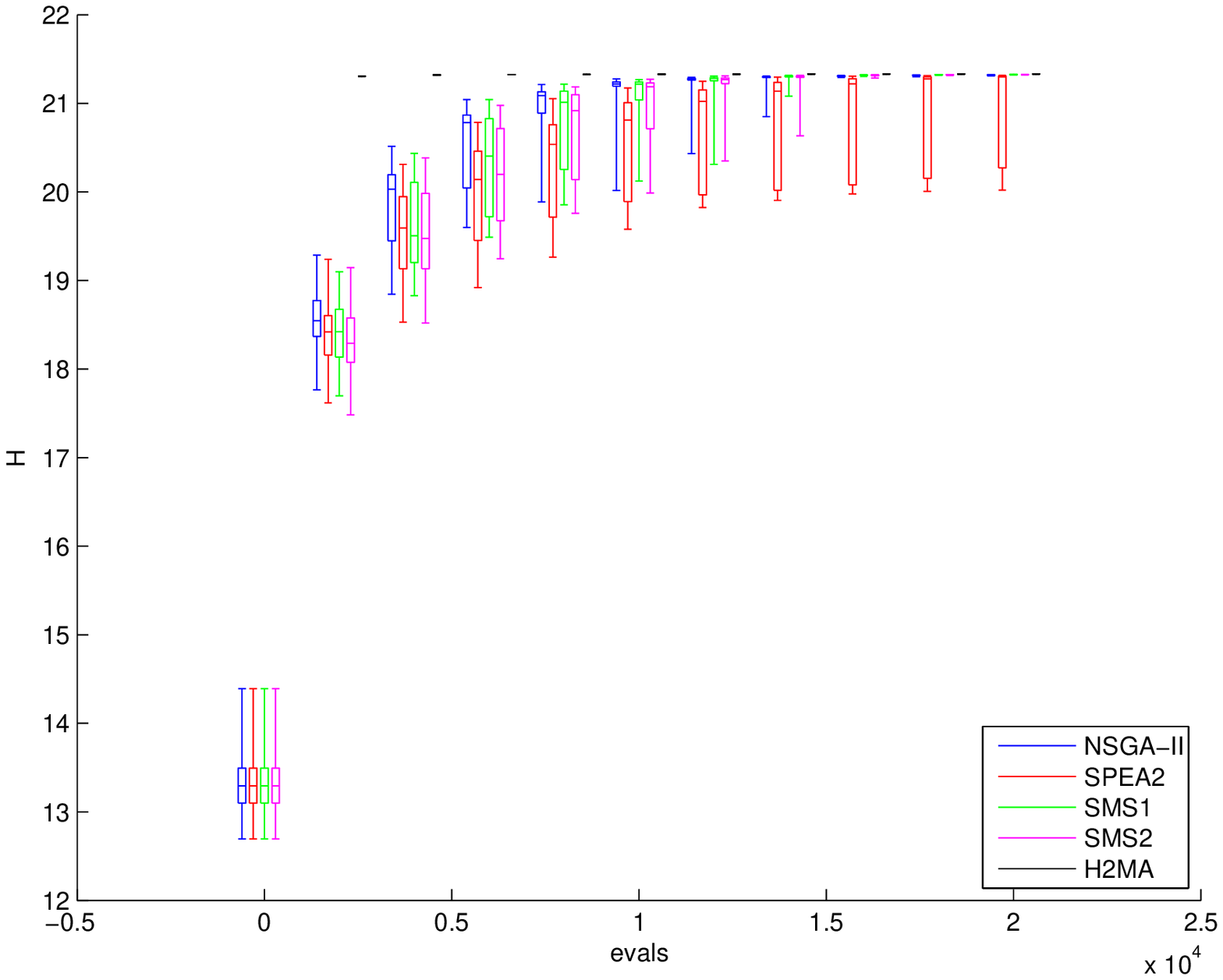}
    \caption{Hypervolume}
    \label{fig:zdt2:H}
  \end{subfigure}
  \begin{subfigure}[b]{0.48\linewidth}
    \psfrag{evals}[c][b]{\# evaluations}
    \psfrag{P}[c][c]{$\log_{10}$ P}
    \includegraphics[width=\linewidth]{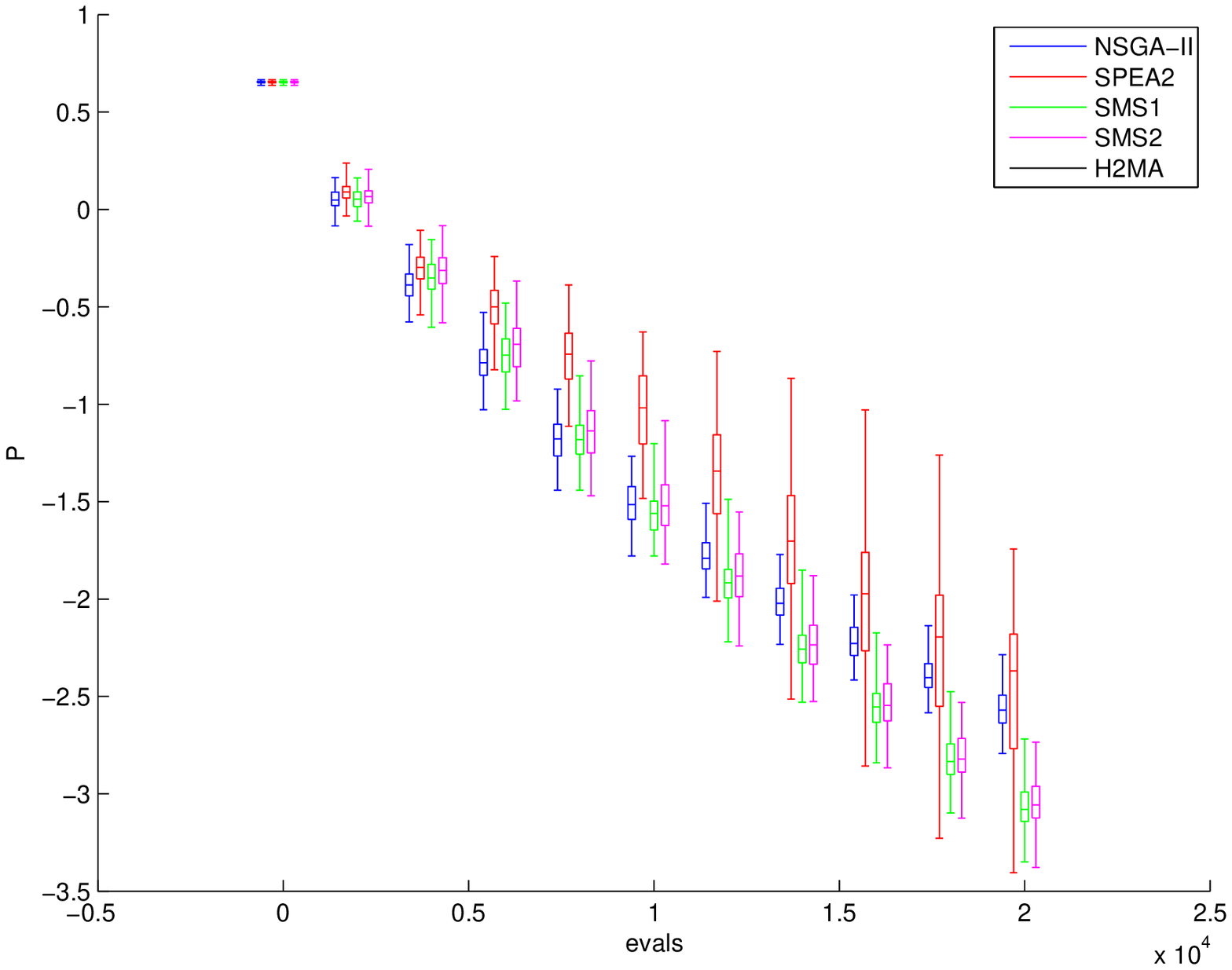}
    \caption{P-distance. Zero values not shown.}
    \label{fig:zdt2:P}
  \end{subfigure}
  \caption{0th, 25th, 50th, 75th, and 100th percentiles every 2000 evaluations
  for the all algorithms on ZDT2.}
  \label{fig:zdt2}
\end{figure*}
\begin{figure*}[t]
  \centering
  \begin{subfigure}[b]{0.48\linewidth}
    \psfrag{evals}[c][b]{\# evaluations}
    \psfrag{H}[c][c]{H}
    \includegraphics[width=\linewidth]{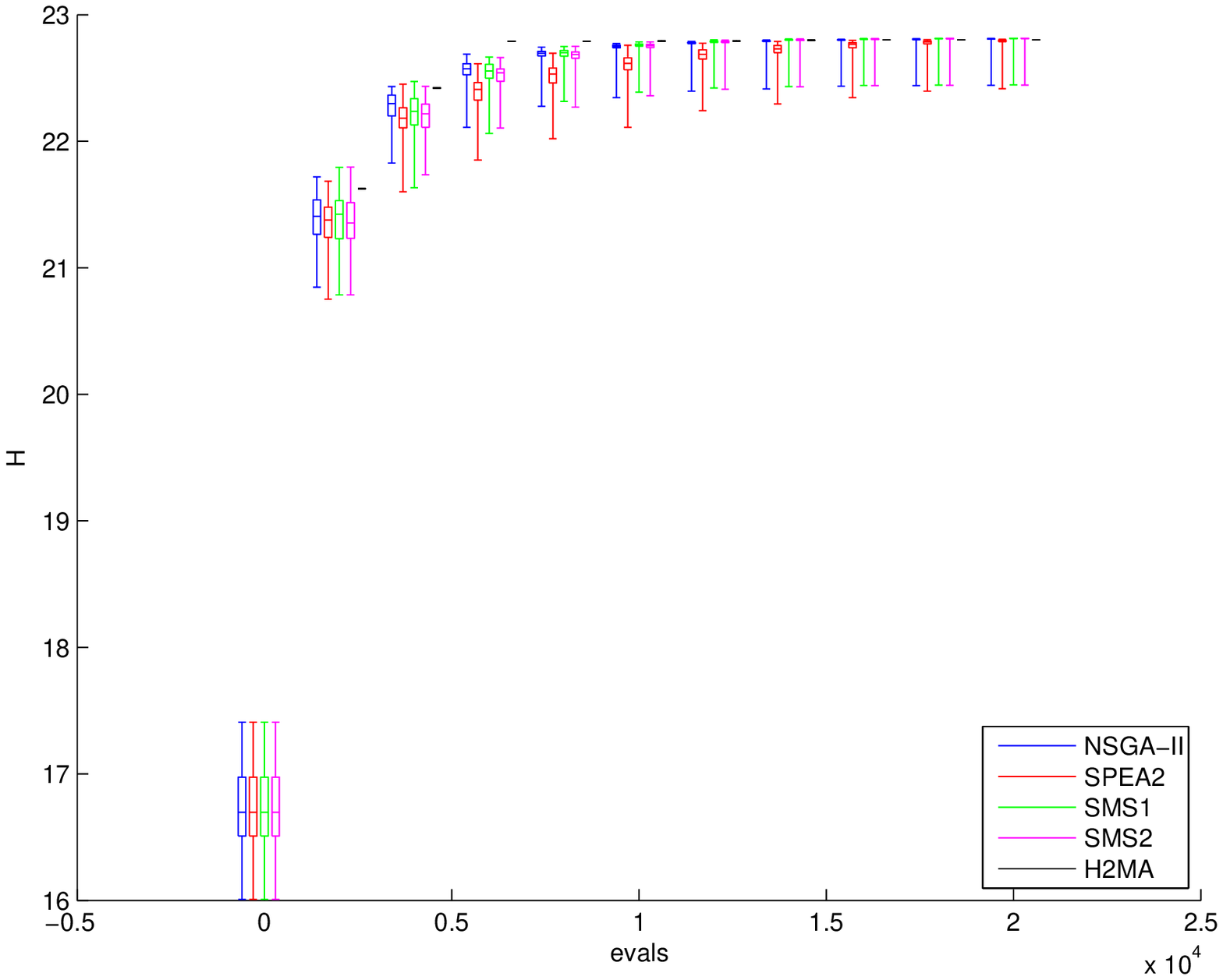}
    \caption{Hypervolume}
    \label{fig:zdt3:H}
  \end{subfigure}
  \begin{subfigure}[b]{0.48\linewidth}
    \psfrag{evals}[c][b]{\# evaluations}
    \psfrag{P}[c][c]{$\log_{10}$ P}
    \includegraphics[width=\linewidth]{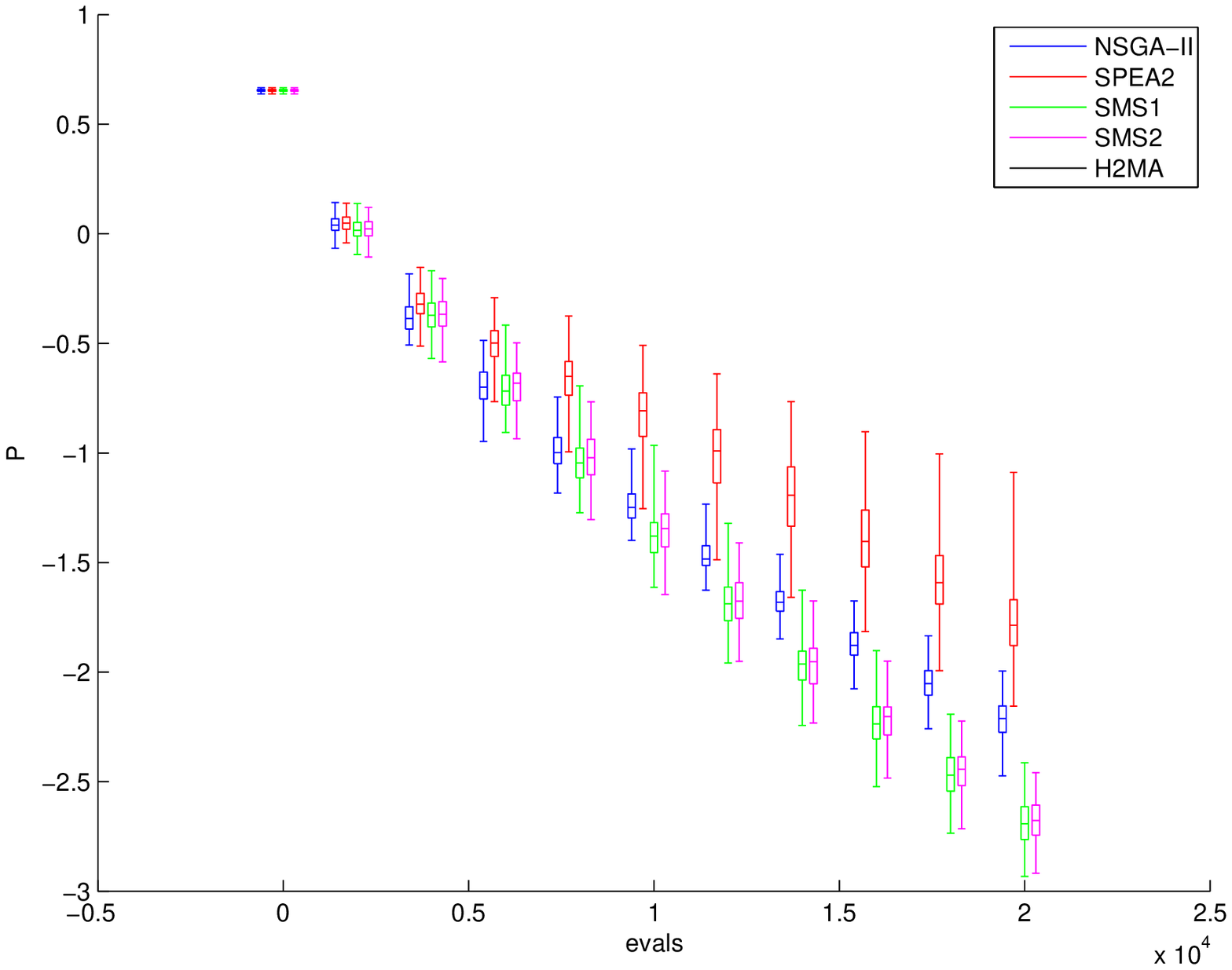}
    \caption{P-distance. Zero values not shown.}
    \label{fig:zdt3:P}
  \end{subfigure}
  \caption{0th, 25th, 50th, 75th, and 100th percentiles every 2000 evaluations
  for the all algorithms on ZDT3.}
  \label{fig:zdt3}
\end{figure*}
\begin{figure*}[t]
  \centering
  \begin{subfigure}[b]{0.48\linewidth}
    \psfrag{evals}[c][b]{\# evaluations}
    \psfrag{H}[c][c]{H}
    \includegraphics[width=\linewidth]{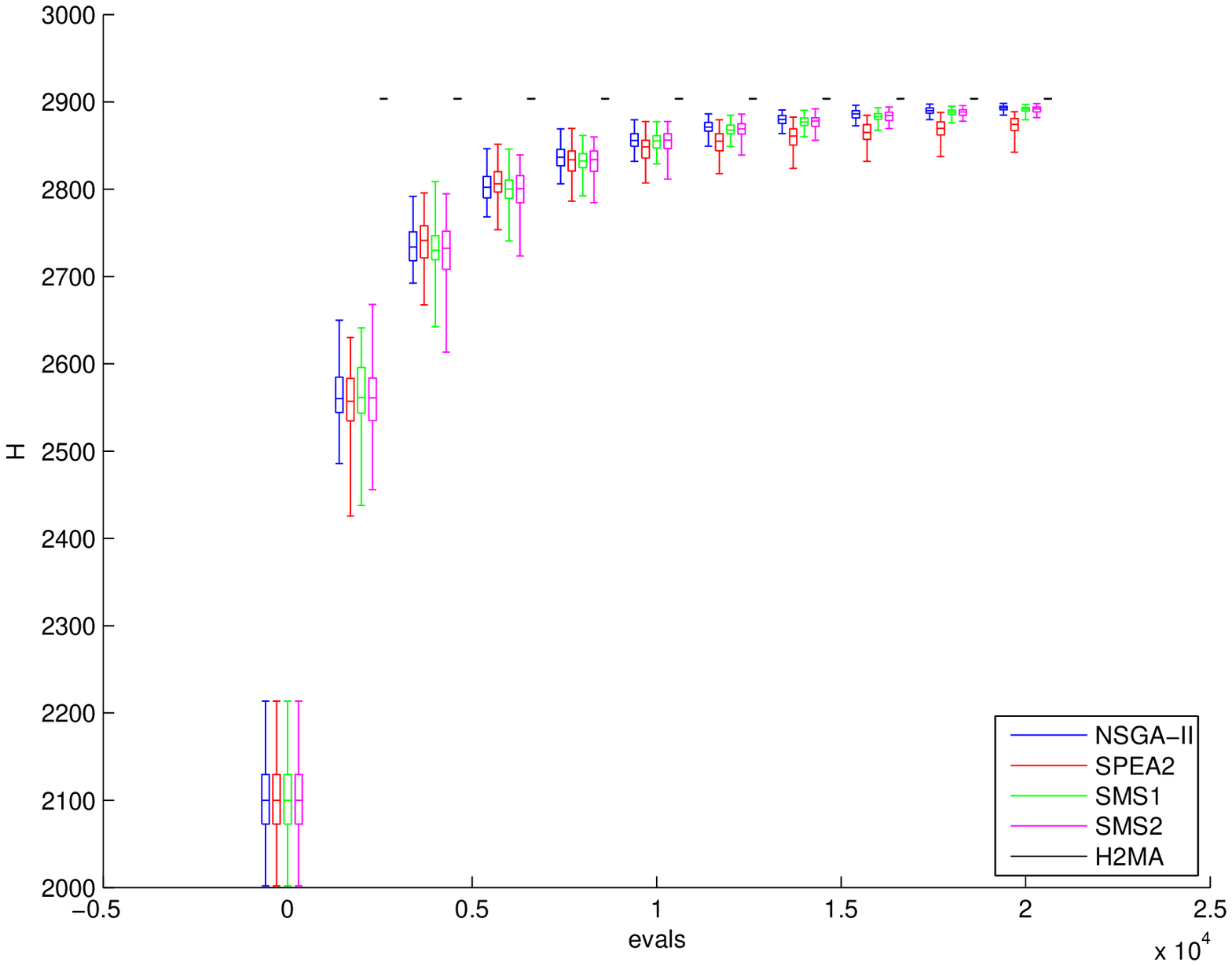}
    \caption{Hypervolume}
    \label{fig:zdt4:H}
  \end{subfigure}
  \begin{subfigure}[b]{0.48\linewidth}
    \psfrag{evals}[c][b]{\# evaluations}
    \psfrag{P}[c][c]{$\log_{10}$ P}
    \includegraphics[width=\linewidth]{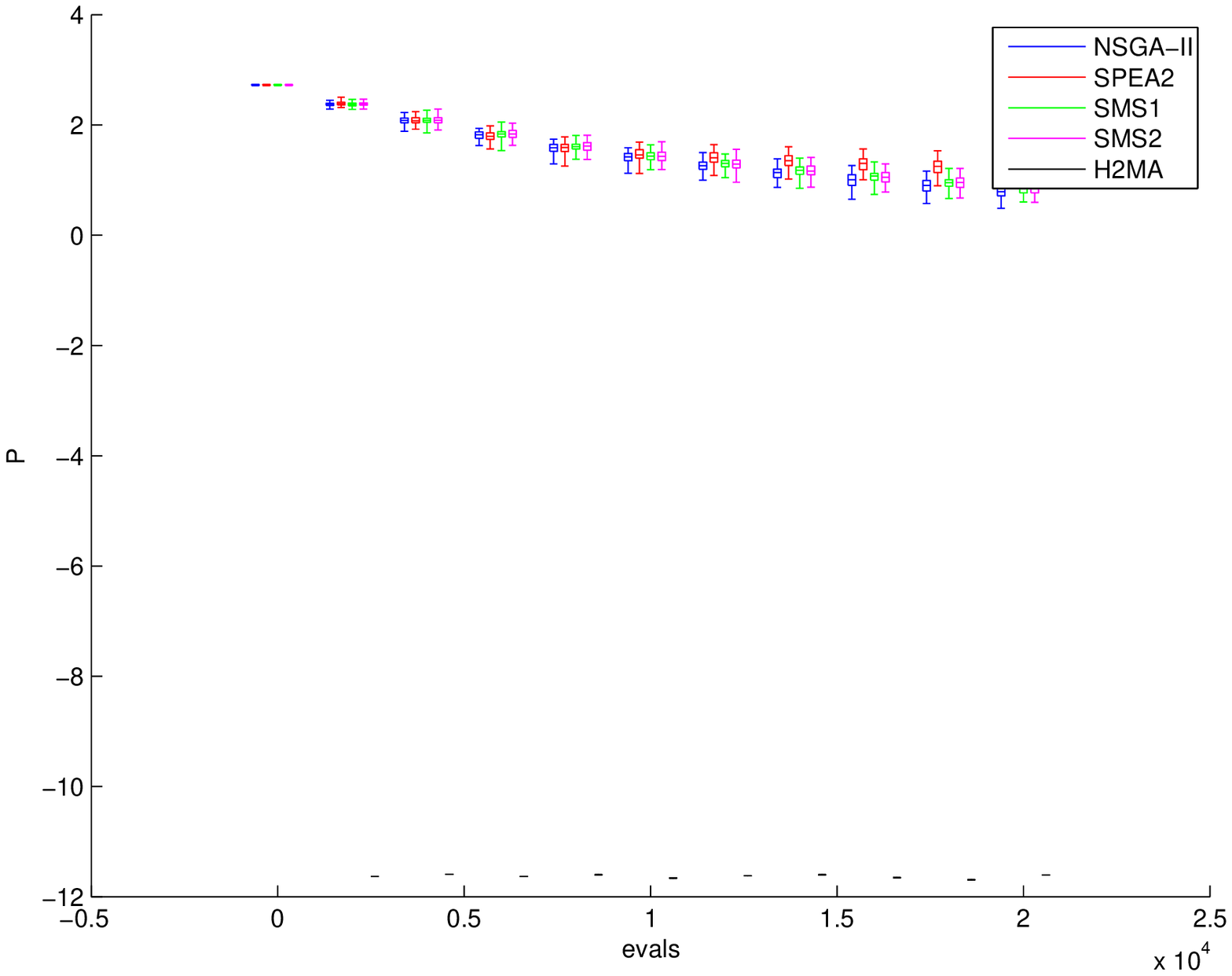}
    \caption{P-distance. Zero values not shown.}
    \label{fig:zdt4:P}
  \end{subfigure}
  \caption{0th, 25th, 50th, 75th, and 100th percentiles every 2000 evaluations
  for the all algorithms on ZDT4.}
  \label{fig:zdt4}
\end{figure*}
\begin{figure*}[t]
  \centering
  \begin{subfigure}[b]{0.48\linewidth}
    \psfrag{evals}[c][b]{\# evaluations}
    \psfrag{H}[c][c]{H}
    \includegraphics[width=\linewidth]{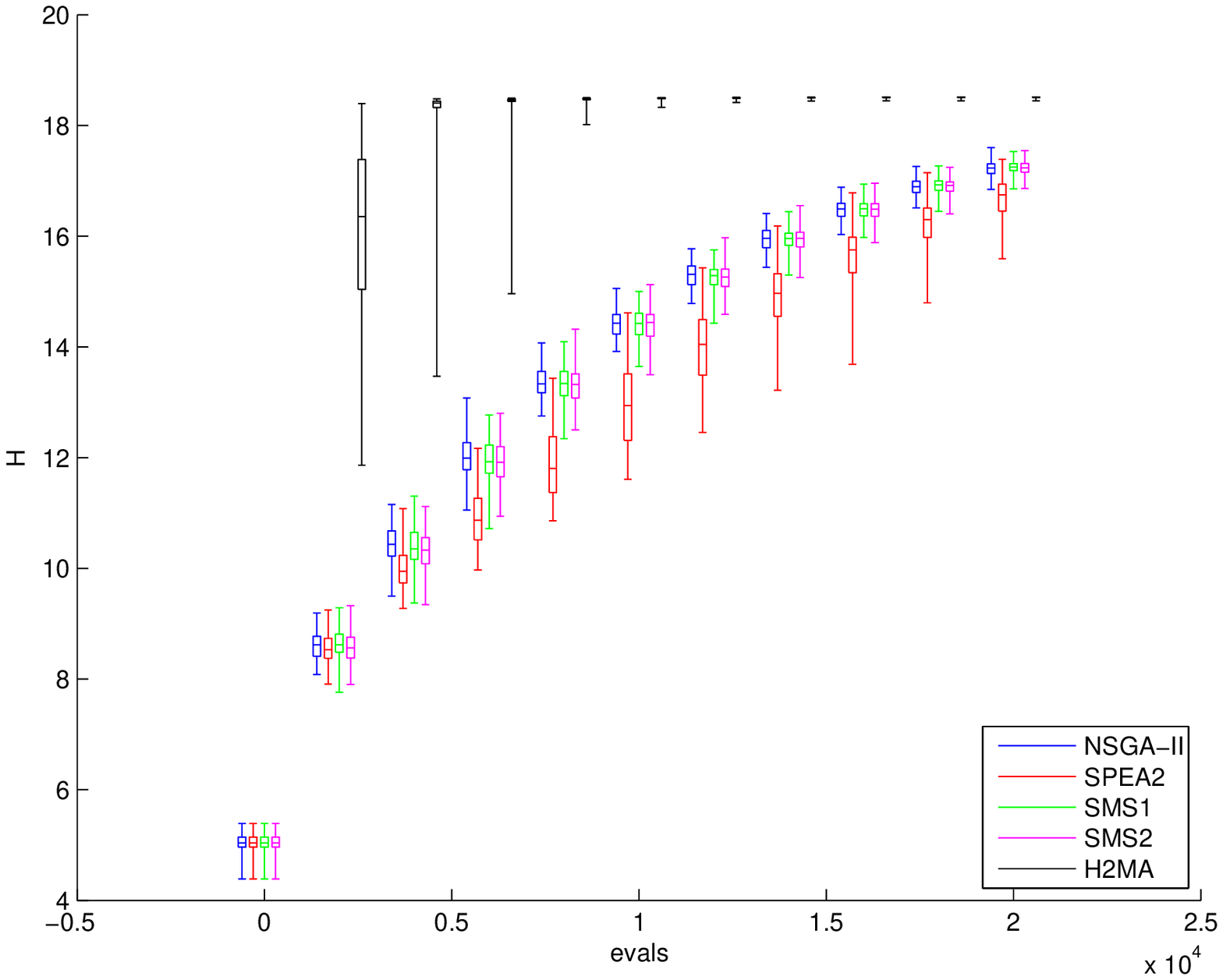}
    \caption{Hypervolume}
    \label{fig:zdt6:H}
  \end{subfigure}
  \begin{subfigure}[b]{0.48\linewidth}
    \psfrag{evals}[c][b]{\# evaluations}
    \psfrag{P}[c][c]{$\log_{10}$ P}
    \includegraphics[width=\linewidth]{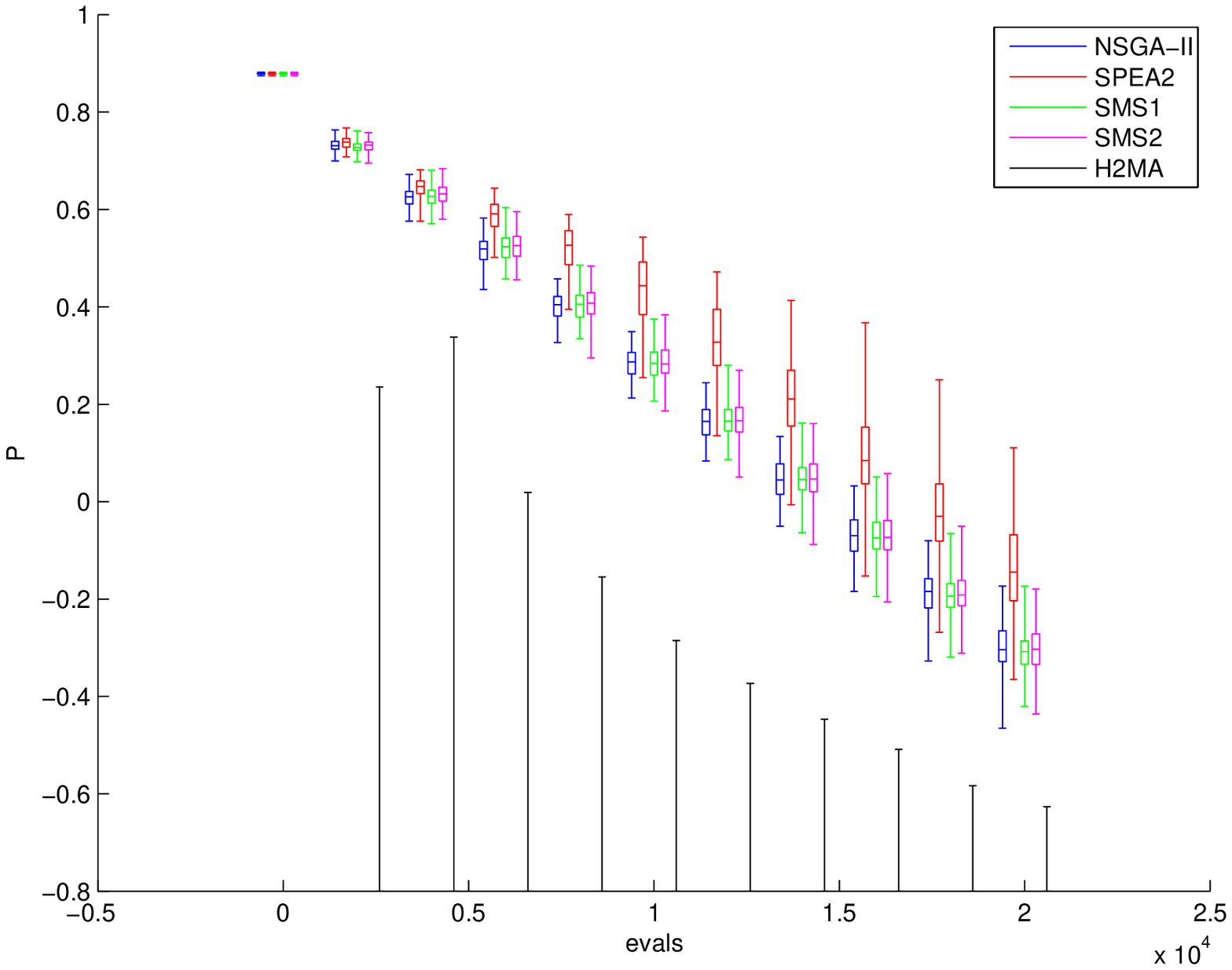}
    \caption{P-distance. Zero values not shown.}
    \label{fig:zdt6:P}
  \end{subfigure}
  \caption{0th, 25th, 50th, 75th, and 100th percentiles every 2000 evaluations
  for the all algorithms on ZDT6.}
  \label{fig:zdt6}
\end{figure*}

\begin{table}[t]
  \centering
  \caption{Benchmark problems used for evaluation. See the Appendix for
  Eqs.~\eqref{eq:zdt1} to~\eqref{eq:zdt6}.}
  \label{tab:experiment}
  \begin{tabular}{|c|c|c|c|}
    \hline
    Problem & Objectives & $\X$ & Nadir point
    \\
    \hline
    ZDT1 & Eq.~\eqref{eq:zdt1} & $[0,1]^n$ & $(2,11)$
    \\
    ZDT2 & Eq.~\eqref{eq:zdt2} & $[0,1]^n$ & $(2,11)$
    \\
    ZDT3 & Eq.~\eqref{eq:zdt3} & $[0,1]^n$ & $(2,11)$
    \\
    ZDT4 & Eq.~\eqref{eq:zdt4} & $[0,1]\times[-5,5]^{n-1}$
    & $(2,2+50(n-1))$
    \\
    ZDT6 & Eq.~\eqref{eq:zdt6} & $[0,1]^n$ & $(2,11)$
    \\
    \hline
  \end{tabular}
\end{table}

We compare our algorithm with existing state-of-the-art multi-objective
optimization algorithms, namely NSGA-II~\cite{deb2002fast},
SPEA2~\cite{zitzler2001spea2}, and SMS-EMOA~\cite{beume2007sms}. All of them
used a population size of 100 individuals. Tests have shown that this size is
able to provide a good performance due to balance between exploration of the
space and exploitation of the individuals, with much less individuals not
providing good exploration and much more not providing good exploitation. The
SMS-EMOA can use two methods for selecting points in dominated fronts: the least
hypervolume contribution or the domination count. Both methods were tested, with
labels SMS1 and SMS2, respectively, in the following figures. Note that this
method only applies for the dominated fronts, since the domination count is zero
for all points in the non-dominated front and the least contributor method must
be used. Furthermore, the SMS-EMOA algorithm's performance presented in this
paper uses a dynamic Nadir point, which is found by adding one to the maximum
over all points in each objective, since using the Nadir points presented in
Table~\ref{tab:experiment} created a very high selective pressure, which in turn
led to poor exploration and performance.

Since the decision space and objectives are continuous, the exploitation and
deterministic exploration methods may resort to a gradient-based algorithm. In
this paper, we used the L-BFGS-B method implemented in the library
SciPy~\cite{scipy}, which is able to handle the bounds of $\X$ and is very
efficient to find a local optimum. As the other algorithms being compared are
evolutionary algorithms, which can only access the objective functions by
evaluating them at given points, the gradient for the L-BFGS-B is computed
numerically to avoid an unfair advantage in favor of our algorithm.

For the stochastic global exploration, we used an evolutionary algorithm with
non-dominance sorting and removal based on the number of dominating points. The
population had a minimum size of $20$ and was filled with the given set of
previous solutions $X$. If less than $20$ points were provided, the others were
created by randomly sampling the decision space $\X$ uniformly. Once a new point
is introduced to the non-dominated front, it is returned for exploitation
because it increases the hypervolume when added to the previous solutions $X$.
The size of this population was chosen experimentally to provide a good enough
exploration of the space toward the initial conditions for the exploitation.
This size is smaller than the population size for the pure evolutionary
algorithms because the pure evolutionary algorithm need diversity to explore and
exploit all of its population, but the stochastic part of the H2MA is already
initialized with good and diverse candidate solutions provided by the
exploitation procedure, reducing its exploration requirements.

Besides computing the solutions' hypervolume, which is the metric that the H2MA
is trying to maximize and that provides a good method for comparing solutions,
we can compute the distance between the achieved objectives and the Pareto
frontier, since the Pareto frontiers for the ZDT functions are known. This
defines the P-distance, which is zero, or close to zero due to numerical issues,
for points at the frontier.

Figs.~\ref{fig:zdt1}, \ref{fig:zdt2}, \ref{fig:zdt3}, \ref{fig:zdt4},
and~\ref{fig:zdt6} present the results for the problems ZDT1, ZDT2, ZDT3, ZDT4,
and ZDT6, respectively. A maximum of 20000 function evaluations was considered,
and the graphs show the $0$th, $25$th, $50$th, $75$th, and $100$th percentiles
for each performance indicator over 100 runs of the algorithms. Since the
P-distance is shown in log-scale, some values obtained by our proposal are
absent or partially present, because they have produced zero P-distance.

From ZDT1 to ZDT4, the H2MA never ran out of regions to explore, so the
stochastic exploration was not used and all runs have the same performance. For
the function ZDT6, the first objective, given by Eq.~\eqref{eq:zdt6:f1}, causes
some problems to the deterministic exploration.

During the creation of the first region for this problem, the mean point is used
as initial condition for optimizing each objective, as shown in
Fig.~\ref{alg:initial}. However, the first objective for ZDT6 has null
derivative when $x_1 = 0.5$. In this case, even traditional higher-order methods
would not help, since the first non-zero derivative of $f_1(x)$ is the sixth. As
the first objective does not change in this case and it also has local minima
that are very hard to overcome, the algorithm quickly switches to using
stochastic exploration. Once new regions have candidate points, the algorithm is
able to exploit them.

Besides this issue in the deterministic exploration of the problem ZDT6, the
local minima of the first objective makes some candidate solutions be
sub-optimal, increasing the P-distance as shown in Fig.~\ref{fig:zdt6:P}.
Nonetheless, the achieved P-distance is better than the evolutionary algorithms
and the $75$th percentile is zero. Moreover, Figs.~\ref{fig:zdt1:P},
\ref{fig:zdt2:P}, \ref{fig:zdt3:P}, and~\ref{fig:zdt4:P} show that the candidate
solutions are always on the Pareto frontier for the problems ZDT1 to ZDT4. This
allows the user to stop the optimization at any number of evaluations, even with
very few function evaluations, and have a reasonable expectation that the
solutions found are efficient.

When we evaluate the hypervolume indicator, we see that, for the problems ZDT1,
ZDT2, ZDT4, and ZDT6, the performance of the H2MA is much better, even for the
last one using stochastic exploration. Moreover, the H2MA's worst hypervolume
was always better than the best hypervolume for all evolutionary algorithms and
it was able to get closer to the maximum hypervolume possible with relatively
few function evaluations, being a strong indication of its efficiency.

For the problem ZDT3, whose hypervolume performance is shown in
Fig.~\ref{fig:zdt3:H}, the H2MA was generally better than the evolutionary
algorithms. The Pareto frontier for ZDT3 is composed of disconnected sets of
points, which was created to test the algorithm's ability to deal with
discontinuous frontiers. Since the exploitation algorithm used for the results
is gradient-based, it is not able to properly handle discontinuous functions,
which is the case of the hypervolume on discontinuous frontiers. However, the
deterministic exploration method is able to find points whose exploitation lay
on the different parts of the Pareto frontier, providing the expected diversity.

\begin{figure}[t]
  \centering
  \psfrag{analytic}[c][b]{Analytic}
  \psfrag{numeric}[c][c]{Numeric}
  \includegraphics[width=\linewidth]{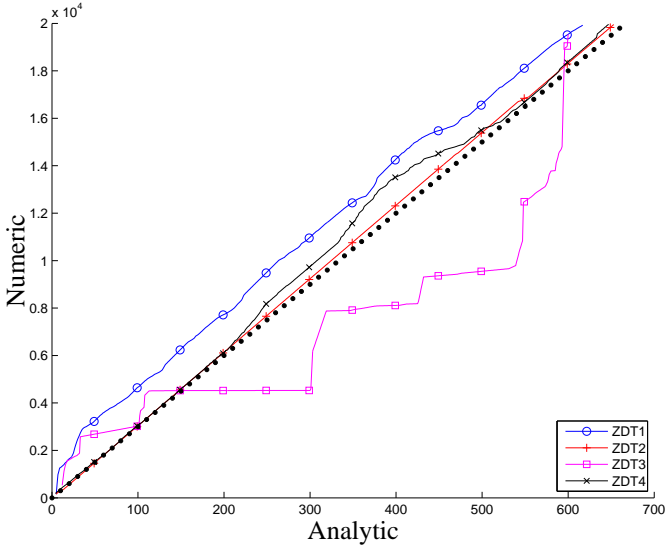}
  \caption{Comparison between the number of function evaluations required to
    achieve the same hypervolume using numeric or analytic gradient. The dotted
  line represents a 30-fold improvement.}
  \label{fig:analytic}
\end{figure}

Fig.~\ref{fig:analytic} shows a comparison between the number of function
evaluations required by the numeric and the analytic gradient to achieve the
same hypervolume on the problems ZDT1 to ZDT4. The analytic method for computing
the hypervolume's gradient is described in~\cite{hypervolume_gradient1}. The
comparison for ZDT6 is not shown due to its different scale, since many function
evaluations are used in the global stochastic exploration because the
deterministic exploration fails to find regions.

As expected, using the analytic gradient causes a 30-fold improvement in
comparison to the numeric gradient, since the number of decision variables is
30. However, the gain is not linear. This can be explained by the difference in
behavior during the deterministic exploration: the first non-dominated point
found is used to perform the exploitation, even if this point was found during
the computation of the numeric gradient. For ZDT1 and ZDT4, this causes the new
points found by the numeric gradient to be very close to the original points,
reducing its performance and increasing the improvement of using the analytic
gradient.

Moreover, a similar effect makes the ZDT3 performance to have a lower
improvement when using the analytic gradient. Since the Pareto frontier for ZDT3
is discontinuous and this causes a discontinuity in the hypervolume, these large
changes can be seen by the numeric gradient because small changes in the
variables can have large effects on the hypervolume, pulling the solution if the
difference is significant, while the analytic gradient is not able to provide
such knowledge. Nonetheless, the analytic gradient presents at least a 15-fold
improvement over the numeric one over the ZDT3.

\subsection{Analysis of the H2MA's performance}
As shown in Section~\ref{sec:results}, the proposed H2MA is able to surpass the
state-of-the-art in multi-objective optimization, based on evolutionary
algorithms. Therefore, it is important to analyze the algorithm and to discuss
why this improvement happened.

Evolutionary algorithms perform a guided exploration, with new individuals
created based on existing high-performing individuals, which allows them to
escape local minima but reduces the convergence speed. On the other hand,
traditional optimization algorithms tend to find local minima quickly, but the
optimal point achieved depends on the minima's regions of attraction.

These two kind of algorithms have complementary natures, which makes them good
candidates for creating a hybrid algorithm: the evolutionary algorithm explores
the space and provides initial conditions for the local optimization, which
then finds minima quickly. Although this does create better results, it only
explains the performance on the ZDT6 problem, since the other problems did not
enter the stochastic phase.

In order to understand the algorithm's behavior, we must keep in mind that each
new point added by the algorithm is solving a very different problem. Since the
previous points that are considered during the hypervolume optimization change
as more points are added, the objective surface for each new point is different
from the previous ones and takes into account the already achieved portion of
the hypervolume. To visualize this, supposed that the hypervolume's gradient is
defined over previously found points and one previous solution is used as the
initial condition for the gradient-based exploitation to find a new point to be
added to the solution set. Although the initial condition was a local optimum
for a previous problem, it is not a local optimum to the current problem,
because any small change that creates a non-dominated point will improve the
total hypervolume. Therefore, we do not need to worry about the new optimization
converging to a previous solution point because the problem landscape is
different and different local minima will be found, increasing the total
hypervolume. The deterministic exploration is only required because the
hypervolume's gradient is not defined at the border of the hypervolume, so a new
independent point must be found.

This explains the performance improvement over ZDT1 to ZDT4, because every added
point improves the hypervolume as much as it can do locally, so that an
improvement is guaranteed to happen. Evolutionary algorithms, on the other hand,
use function evaluations without guarantees of improvement of the total
hypervolume, since dominated solutions can be found.

Moreover, although a local optimum found during exploitation may not be an
efficient solution due to irregularities in the objective surface, the
experiments show that this is not the case most of the time, since the
P-distance of the solutions found are generally zero. This result is expected,
since the hypervolume is maximal when computed over points in the Pareto set,
and the performance on all ZDT problems provide support to this claim.

We must highlight that we are not saying that evolutionary algorithms should not
be used at all, but that they should be applied whenever traditional
optimization methods are not able to solve the problem. This is the case of the
ZDT6, for instance, where an evolutionary algorithm was required to provide
initial conditions for the exploitation. We consider very important to have
alternative methods that are better on a subset of the problems and to use them
when a problem from such subset is present. This is exactly what the H2MA does:
when the traditional optimization is not able to find an answer, which indicates
that the problem is outside of the subset with which it can deal, an
evolutionary algorithm, which is able to handle a superset class of problems, is
used until the problem becomes part of the subset again, establishing a
switching behavior that takes advantage of both algorithms.
\section{Conclusion}
\label{sec:conclusion}
This paper proposed the Hybrid Hypervolume Maximization Algorithm (H2MA) for
multi-objective optimization, which tries to maximize the hypervolume one point
at a time. It first tries to perform deterministic local exploration and, when
it gets stuck, it switches to stochastic global exploration using an
evolutionary algorithm. The optimization algorithm used during deterministic
optimization is problem-dependent and can be given by a gradient-based method,
when the decision space is continuous, or a hill-climbing method, when the
decision space is discrete. Here we have explored solely continuous decision
spaces.

The algorithm was compared with state-of-the-art algorithms for multi-objective
optimization, namely NSGA-II, SPEA2, and SMS-EMOA on the ZDT problems. Despite
using numeric gradient for the objective functions, which increases the number
of function calls, the algorithm consistently provided a higher hypervolume for
the same number of function evaluations when compared to the aforementioned
evolutionary algorithms. Only for the ZDT3 the performance was slightly reduced
due to the discontinuous nature of the Pareto frontier, which causes a
discontinuity in the hypervolume, not properly handled by gradient-based
methods.

Moreover, for all problems except for ZDT6, all the solutions found by the
algorithm were over the Pareto frontier, which makes them efficient solutions.
For the ZDT6, the median case also had all solutions over the Pareto frontier,
but the use of the stochastic exploration not always guided to a solution at the
Pareto frontier. Nonetheless, the obtained solutions were better than those
provided by the evolutionary algorithms. Moreover, the solutions provided for
ZDT1 to ZDT4 achieved high performance using only the deterministic part of the
algorithm.

Evolutionary algorithms usually have better performance when their populations
are larger, so that diverse individuals can be selected for crossover. However,
most of the time people do not require many options, so the H2MA presents itself
as an alternative choice for finding a good set of solutions at a lower
computational cost in most problems, although it does not limit the
computational burden and the number of points found. If the problem has more
reasonable objectives than ZDT6, which was designed with an extreme case in
mind, we can expect that many points will be found by the deterministic
mechanisms, which makes the algorithm more reliable. Moreover, the solutions
found should be efficient, which is characterized by a low P-distance, and
diverse on the objectives, which is characterized by a larger hypervolume when
only efficient solutions are considered.

Future work should focus on using surrogates to reduce the number of
evaluations~\cite{jin2005comprehensive,knowles2008meta,voutchkov2010multi}.
Although the H2MA is very efficient on its evaluations, the numeric gradient may
consume lots of evaluations and be unreliable for complicated functions, as
their implementation can cause numerical errors larger than the step used. Using
a surrogate, the gradient can be determined directly and less evaluations are
required.

Another important research problem is to find a new algorithm for computing the
hypervolume, since existing algorithms are mainly focused on computing the
hypervolume given a set of points~\cite{beume2009complexity}. Since the solution
set is constructed one solution at a time in the H2MA, a recursive algorithm
that computes the hypervolume of $X \cup \{x\}$ given the hypervolume of $X$
should reduce the computing requirement.

\appendix
ZDT1:
\begin{subequations}
\label{eq:zdt1}
\begin{align}
  f_1(x) &= x_1
  \\
  f_2(x) &= g(x)h(f_1(x),g(x))
  \\
  g(x) &= 1 + \frac{9}{n-1} \sum_{i=2}^{n} x_i
  \\
  h(f_1(x),g(x)) &= 1-\sqrt{f_1(x)/g(x)}
\end{align}
\end{subequations}

ZDT2:
\begin{subequations}
\label{eq:zdt2}
\begin{align}
  f_1(x) &= x_1
  \\
  f_2(x) &= g(x)h(f_1(x),g(x))
  \\
  g(x) &= 1 + \frac{9}{n-1} \sum_{i=2}^{n} x_i
  \\
  h(f_1(x),g(x)) &= 1-(f_1(x)/g(x))^2
\end{align}
\end{subequations}

ZDT3:
\begin{subequations}
\label{eq:zdt3}
\begin{align}
  f_1(x) &= x_1
  \\
  f_2(x) &= g(x)h(f_1(x),g(x))
  \\
  g(x) &= 1 + \frac{9}{n-1} \sum_{i=2}^{n} x_i
  \\
  h(f_1(x),g(x)) &= 1-\sqrt{\frac{f_1(x)}{g(x)}} - \sin(10\pi
  f_1(x))\frac{f_1(x)}{g(x)}
\end{align}
\end{subequations}

ZDT4:
\begin{subequations}
\label{eq:zdt4}
\begin{align}
  f_1(x) &= x_1
  \\
  f_2(x) &= g(x)h(f_1(x),g(x))
  \\
  g(x) &= 1 + 10(n-1) + \sum_{i=2}^{n} (x_i^2 - 10 \cos(4\pi x_i))
  \\
  h(f_1(x),g(x)) &= 1-\sqrt{f_1(x)/g(x)}
\end{align}
\end{subequations}

ZDT6:
\begin{subequations}
\label{eq:zdt6}
\begin{align}
  f_1(x) &= 1-\exp(-4x_1) \sin^6(6\pi x_1)
  \label{eq:zdt6:f1}
  \\
  f_2(x) &= g(x)h(f_1(x),g(x))
  \\
  g(x) &= 1 + 9 \left(\sum_{i=2}^{n} \frac{x_i}{n-1}\right)^{0.25}
  \\
  h(f_1(x),g(x)) &= 1-(f_1(x)/g(x))^2
\end{align}
\end{subequations}

\section*{Acknowledgment}
The authors would like to thank CNPq for the financial support.

\bibliographystyle{templates/IEEEtran/bib/IEEEtran}
\bibliography{paper}

\begin{IEEEbiographynophoto}{Conrado S. Miranda}
  received his M.S. degree on Mechanical Engineering and his B.S. in Control
  and Automation Engineering from the University of Campinas (Unicamp), Brazil,
  in 2014 and 2011, respectively. He is currently a Ph.D. student at the School
  of Electrical and Computer Engineering, Unicamp. His main research interests
  are machine learning, multi-objective optimization, neural networks, and
  statistical models.
\end{IEEEbiographynophoto}

\begin{IEEEbiography}[{\includegraphics[width=1in,height=1.25in,clip,keepaspectratio]{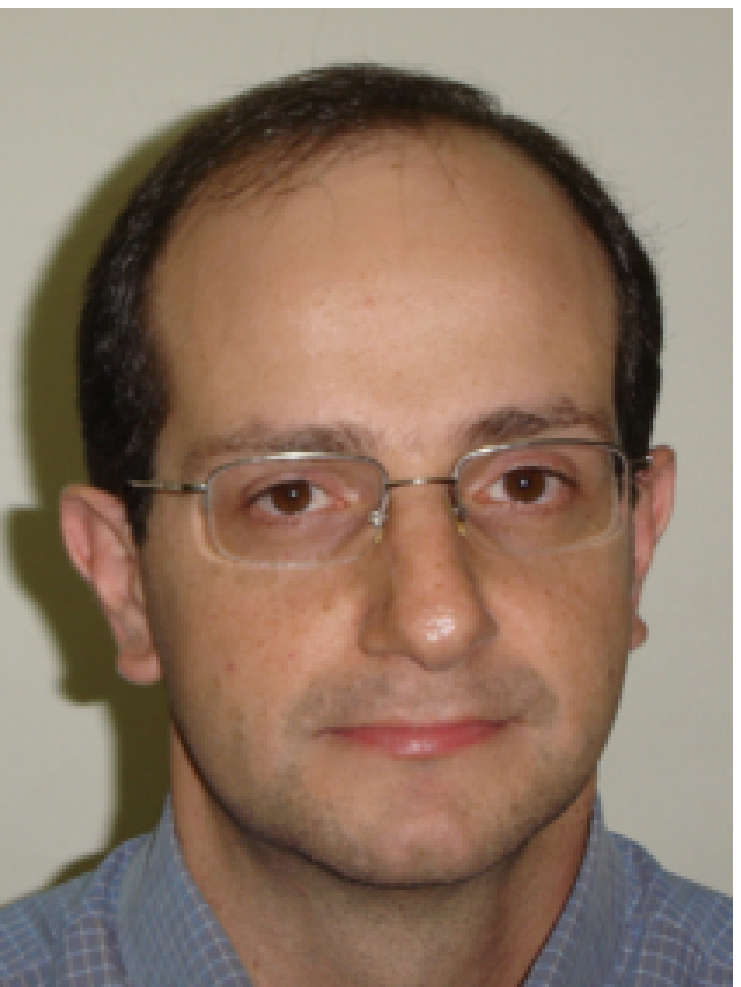}}]{Fernando J. Von Zuben}
  received his Dr.E.E. degree from the University of Campinas (Unicamp),
  Campinas, SP, Brazil, in 1996. He is currently the header of the Laboratory of
  Bioinformatics and Bioinspired Computing (LBiC), and a Full Professor at the
  Department of Computer Engineering and Industrial Automation, School of
  Electrical and Computer Engineering, University of Campinas (Unicamp). The
  main topics of his research are computational intelligence, natural computing,
  multivariate data analysis, and machine learning. He coordinates open-ended
  research projects in these topics, tackling real-world problems in the areas
  of information technology, decision-making, pattern recognition, and discrete
  and continuous optimization.
\end{IEEEbiography}

\vfill

\end{document}